\RequirePackage{amsmath}
\documentclass[runningheads]{llncs}
\usepackage[T1]{fontenc}
\usepackage{graphicx}
\usepackage{booktabs}
\usepackage[misc]{ifsym}
\newcommand{\corr}{(\Letter)}
\usepackage{graphicx}
\usepackage{multirow}
\usepackage{adjustbox}
\usepackage{subcaption}
\usepackage{verbatim}
\usepackage{amssymb}

\usepackage{float}
\floatstyle{plaintop}
\restylefloat{table}

\usepackage{cite}

\begin{document}
\title{Model Fusion via Neuron Transplantation
}
%
%

\author{Muhammed Öz\orcidID{0009-0003-4337-407X} 
 \and \\
Nicholas Kiefer\orcidID{0009-0008-1454-4144}
\and \\
Charlotte Debus\orcidID{0000-0002-7156-2022} 
\and \\
Jasmin Hörter 
\and \\
Achim Streit\orcidID{0000-0002-5065-469X} 
\and \\
Markus Götz\orcidID{0000-0002-2233-1041}\corr
}

\tocauthor{Muhammed~Öz,Nicholar~Kiefer,Charlotte~Debus,Jasmin~Hörter,Achim~Streit,Markus~Götz}
\toctitle{Model Fusion via Neuron Transplantation}

%

\authorrunning{M. Öz et al.}

%






\institute{Scientific Computing Center (SCC), Karlsruhe Institute of Technology (KIT),\\
Hermann-von-Helmholtz Platz 1, 76344 Eggenstein-Leopoldshafen, Germany \\
\email{\{muhammed.oez, nicholas.kiefer, charlotte.debus, jasmin.hoerter, achim.streit, markus.goetz\}@kit.edu}
}

\maketitle              
%


\begin{abstract}

Ensemble learning is a widespread technique to improve the prediction performance of neural networks. However, it comes at the price of increased memory and inference time. In this work we propose a novel model fusion technique called \emph{Neuron Transplantation (NT)} in which we fuse an ensemble of models by transplanting important neurons from all ensemble members into the vacant space obtained by pruning insignificant neurons. An initial loss in performance post-transplantation can be quickly recovered via fine-tuning, consistently outperforming individual ensemble members of the same model capacity and architecture. Furthermore, NT enables all the ensemble members to be jointly pruned and jointly trained in a combined model. Comparing it to alignment-based averaging (like Optimal-Transport-fusion), it requires less fine-tuning than the corresponding OT-fused model, the fusion itself is faster and requires less memory, while the resulting model performance is comparable or better. The code is available under the following link: \\
\url{https://github.com/masterbaer/neuron-transplantation} .

\end{abstract}
\section{Introduction}

Nowadays, it is not uncommon to have neural networks with billions of parameters~\cite{chowdhery2022palm}. The inference time and memory demand for large pretrained models is costly, even more so when the model is only one of many base learners in an ensemble. Ensemble learning techniques are known to increase generalization~\cite{MOHAMMED2023757}, but require even more memory and computational resources compared to a single model. This makes the deployment of deep ensembles difficult, if not even unattainable~\cite{DBLP:journals/corr/abs-2101-08387}. Knowledge distillation~\cite{hinton2015distilling} is one solution to compress ensembles but it requires the ensemble members in the fine-tuning process. An alternative to this is weight-based model fusion such as weight averaging and neuron alignment~\cite{singh2023model}. Yet both come with an issue: weight averaging can lead to a poor performance as the different models can lie in different basins in the loss landscape~\cite{ainsworth2023git} and are therefore separated by a so-called loss barrier. Meanwhile alignment based methods, e.g. graph matching~\cite{pmlr-v162-liu22k} or layerwise neuron alignment~\cite{singh2023model}, usually come with large memory demands~\cite{li2023deep}. Hence we focus on a less demanding compression method: pruning.

The study of pruning in neural networks has shown that not all parameters contribute equally to the predictive performance~\cite{liebenwein2021lost} and may be removed without any or negligible impact. While ensemble-pruning usually refers to the selection of a representative subset of the ensemble~\cite{10.5555/2888116.2888125}, we create a pruning-based model fusion technique called \emph{Neuron Transplantation (NT)} which fuses the ensemble members by transplanting only their most important neurons into the fused model. Since we avoid weight averaging, we also avoid the accompanying issue of loss barriers. We show that transplantation captures the essence of an ensemble without significant losses, as the ensemble performance can be partially recovered via fine-tuning compensating the loss of the smaller neurons.

Neuron Transplantation can be used in place of (aligned) weight averaging for diverse models. Prime examples are federated learning~\cite{QI2024272}, deep ensemble pruning~\cite{dongSurveyEnsembleLearning2020}, fine-tuning of pretrained models for downstream tasks~\cite{wortsman2022model}, knowledge distillation as a way to fuse an ensemble teacher or an ensemble student~\cite{singh2023model}, or possibly for synchronous SGD training~\cite{sun2017ensemblecompression, das2016distributed}. It should not be used for too similar models as the transplanted neurons are redundant.

Our contributions can be summarized as follows: 
\begin{enumerate}
    \item We present a novel ensemble-compression technique which fuses models of the same architecture into a same sized one, with minimal compression loss mitigated through fine-tuning outperforming state-of-the-art model fusion techniques.
    \item We find that our approach is susceptible to the order of operations, first merging, then jointly pruning and fine-tuning the neurons leads to slightly better results than individually pruning, merging and fine-tuning. Thus we recommend to first merge when using NT for model fusion. For ensemble pruning, the merging step can be omitted as it introduces cross-weights.
    \item We find that merging multiple models jointly or in a hierarchical way leads to the best results while an iterative approach---due to unequal weighting---leads to slightly worse performance recovery. 
    \item We show through various ablations with different widths, depths and number of models, that NT is applicable in all settings while the fusion results in diminishing returns for an increasing amount of models. 
    \item We find and discuss a fundamental limitation of NT: fusing a model with itself (and very similar models) replaces the smaller neurons with already known ones which leads to an overall information loss without the compensation of new neurons. This limits NT to merging models that are ``diverse enough''. 
\end{enumerate}

\section{Related Work}

The field of model fusion has been categorized by Li et al.~\cite{li2023deep} into four categories, of which ``weight averaging'', and ``alignment'' is of relevance to our work.

\paragraph{Weight Averaging.}
Weight averaging of trained models~\cite{DBLP:journals/corr/abs-2002-06440} is the process of combining models in a defined space. The outputs of the individual models are weighted and summed. Weight averaging has the problem of a loss barrier which increases the loss for fused models through averaging. For large pre-trained models, it has been shown that they lie in a single low error basin~\cite{DBLP:journals/corr/abs-2008-11687, wortsman2022model}, rendering vanilla averaging feasible. Another such scenario is the case of merging similar models~\cite{li2023deep}. In such fusion settings~\cite{inproceedings, Leontev_2019}, no loss-barrier seems to be present making average-based methods viable.
Unlike the approaches detailed here, we do not interpolate between different models, but instead transplant all of them into the fused model to create a new initialization that is able to reach near-ensemble performance with the memory and speed of a single member.

\paragraph{Alignment.} 
Entezari et al. conjecture that models can be permuted in such a way that there is no loss barrier on the linear interpolation between them~\cite{entezari2022role}. This solves the problem of the loss barrier in weight averaging, but the search space to probe all possible permutations in a reasonable time is too large. To solve this, Ainsworth et al. propose activation matching, weight matching and using a straight-through estimator to find good permutations~\cite{ainsworth2023git}. Singh and Jaggi use Optimal Transport (OT) to align the neurons (layerwise, in a greedy way) minimizing the total transportation cost between them~\cite{singh2023model}. In both works, a certain loss barrier remains when the actual averaging is done, especially when layers with small widths are fused. In contrast to these methods, we do not try to ``align and average'' neurons, but to ``select'' the most relevant ones from the ensemble members, hence avoiding the averaging-induced loss.



\paragraph{Distillation.}
Knowledge distillation~\cite{hinton2015distilling} is a compression method that can be used as a fusion method~\cite{DBLP:journals/corr/abs-2011-07449, DBLP:journals/corr/abs-1812-02425, DBLP:journals/corr/abs-2012-09816}. It is orthogonal to averaging or transplanting and can be used alongside other fusion methods such as OT~\cite{singh2023model} or ours. In knowledge distillation, knowledge from the teacher is distilled into the student by having access to the logits of the teacher model as soft targets, in addition to the original labels as hard targets. Sun et al. use knowledge distillation on a diverse ensemble as a synchronization method for SGD~\cite{sun2017ensemblecompression}. After each fusion, workers train independently until the next synchronization step. To ensure diversity of the ensemble members, they add a similarity term to the loss function. We use distillation to further study the behaviour of our method in experiments.

\paragraph{Pruning.}
Pruning is the method of removing elements from a model, the smallest unit being a single weight, which can be set to zero to decrease model complexity while retaining predictive performance. The field of pruning is large~\cite{mishra2020survey}. Structured pruning considers compositions of elements to be pruned concurrently, for linear layers these might be neurons, their equivalent for CNNs might be filters~\cite{10.1145/3005348}. Units to be removed are selected by some measure, e.g. magnitude pruning uses the $L_p$-norm as a simple indicator whether a units contribution is of relevance~\cite{248452}. As structured pruning can get complex when using different architectures, we resort to \emph{torch-pruning}~\cite{fang2023depgraph} which is a pruning framework that can handle many structural couplings in a general way.



\section{Neuron Transplantation}


\begin{figure}
  \makebox[\textwidth][c]
  {\includegraphics[width=0.5\linewidth]{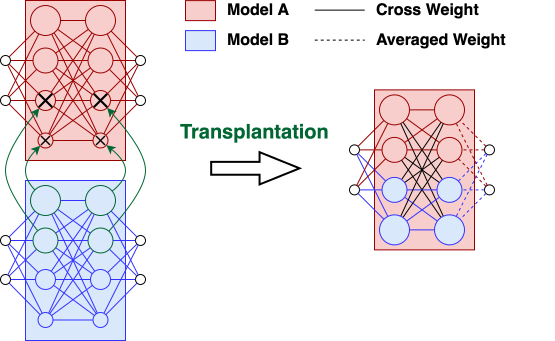}}
  \caption{Neuron Transplantation. Low-magnitude neurons are replaced by large-magnitude ones from other models.}
  \label{fig:Neuron Transplantation_process}
\end{figure}

\begin{figure}
  \makebox[\textwidth][c]
  {\includegraphics[width=1.0\linewidth]{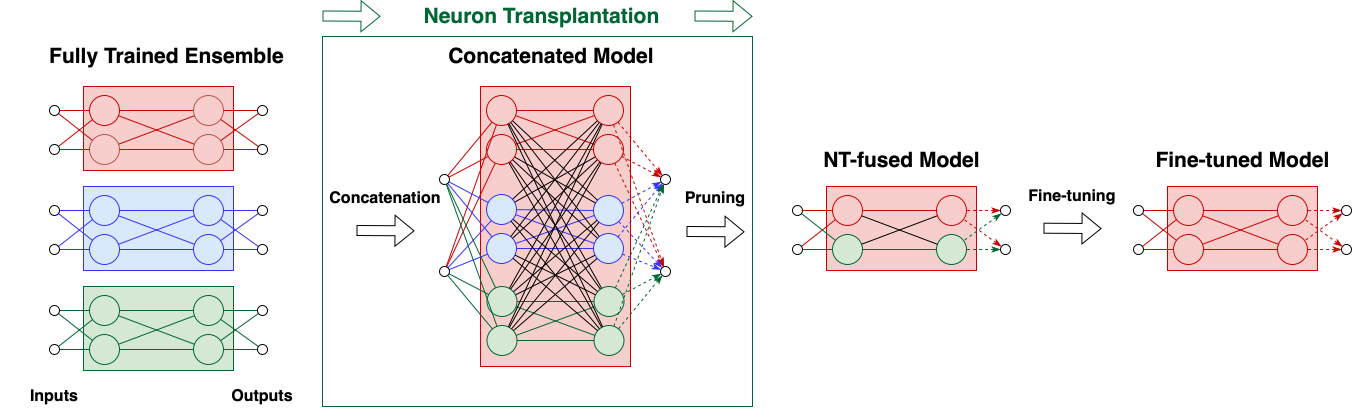}}
  \caption{Pipeline of fusing multiple ensemble members. Multiple models are trained independently, concatenated into one large model, pruned down to the original size and then fine-tuned.}
  \label{fig:process}
\end{figure}

Our proposed \emph{Neuron Transplantation} method (see Figure \ref{fig:process}) is an ensemble fusion method that merges important neurons in a layer-wise fashion. We do not try to ``align and average'' neurons, but to ``select'' the most relevant ones from each ensemble member. Consider an ensemble which consists of $k$ members with equal architecture and equal size of all layers and parameters. \textit{NT} is the composition of the following steps:
\begin{enumerate}
    \item Initialize all models with different random seeds, and train individually using the full data set.
    \item Concatenate the non-output layers vertically, and average the classification layers.
    \item Prune all non-output layers via structured magnitude pruning to a sparsity of $1-\frac{1}{k}$ to obtain the original architecture of a single model.
    \item Using the new initialization obtained from the steps above, fine-tune the resulting model on the full data set to compensate for the lost smaller magnitude neurons.
\end{enumerate}
 
\paragraph{Concatenation.}
Step two of our method is done in a way that the resulting model is equivalent to output averaging. Layers connected to the input stay connected to the input, the output layer is an average of all individual output layers and all others are concatenated. All weights between the layers, which we call \textit{cross weights}, are initialized to zero, and will later be learned by fine-tuning the model. The concatenation is illustrated in Figure \ref{fig:process}.

Formally, for an input $x \in \mathbb{R}^n$ a linear layer is of the form $Wx + b$ with weight matrix $W \in \mathbb{R}^{m \times n}$ and bias $b \in \mathbb{R}^{m}$, $n$, $m$ being the input and output dimension respectively. The fusion of linear layers splits into three distinct cases. The matrix resulting from the concatenation of input layers has shape $\mathbb{R}^{km \times n}$ with $k$ the number of models in the ensemble. Any layer not connected to the input or output will have shape $\mathbb{R}^{km \times kn}$. The output layer is an average of all individual output layers and has shape $\mathbb{R}^{m \times kn}$.

\begin{figure}[htbp]
 \includegraphics[width=1\textwidth]{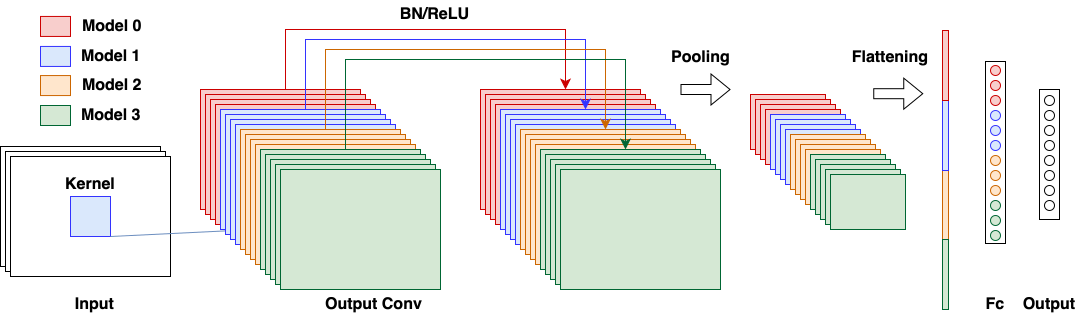}
 \caption{Concatenating 2D convolution layers. Channels are stacked,  batch normalization and pooling operations are preserved.}
 \label{fig:cnn}
 \hfill
\end{figure}

Analogously, a concatenation operation can be defined for (2D-) convolutions and operations that usually are used along convolutions such as batch normalizations and poolings. Figure \ref{fig:cnn} illustrates the different operations when concatenating convolutional layers.

The weight matrix $W \in \mathbb{R}^{o \times i, \times h \times w}$ and bias $b \in \mathbb{R}^{o \times h \times w}$, of a 2D convolution, where $o, i, h, w$ are the output, input, kernel height and kernel width dimensions respectively, can be combined with equally sized weight matrices of other layers, by stacking them along the channel dimension. Because we do not apply convolutional layers as output layers, we do not have to handle this case.

For Convolutional Neural Networks (CNNs), operations such as Batch Normalizations~\cite{DBLP:journals/corr/IoffeS15}, pooling and flattening are usually used in conjunction. Batch-Normalizations only act on single channels. Since the layers are concatenated channel-wise, the parameters of the fused model (the weight, the bias, the running mean and the running variance) are simple concatenations of the individual models parameters. Poolings and flattenings need no special consideration. 


\paragraph{Pruning.}
In step three, the fused model is pruned. In order to obtain the architecture and size of an individual model, we use structured magnitude-pruning to prune nodes. Note that it is also possible to obtain other sizes as in Figure \ref{fig:interpolation} where larger sizes recover more of the ensemble performance and smaller sizes less. For this work we stick to the original architecture. In a layer-wise fashion, the neurons with the smallest $L_2$-norm are removed. We chose the $L_2$-norm due to its simplicity, but we expect similar performance when pruning with other metrics such as the $L_1$-norm~\cite{li2017pruning}. We do this until the original architecture is obtained, i.e. for k models, we use a sparsity of $1-1/k$.

\subsection{Analysis of NT}
\label{propertydiscussion}

\paragraph{Memory and Time Requirements.}
Let the ensemble consist of $k$ equal sized models. For parallel training and joint fusion, all individual models need to be stored in memory. Storing the $L_2$-norms for each neuron is the only additional memory needed.

For time constraints, one only needs to consider the upper bound of sorting the computed $L_2$-norms of each neuron. For a layer with $k*N$ neurons and $M$ inputs each, computing the $L_2$-norms is of order $\mathcal{O}(kNM)$ and sorting them an additional $\mathcal{O}(kN log(kN))$. This is negligible compared to training and fine-tuning, especially if $k$ is small. For very large $k$, it may be beneficial to use a reduction scheme for the fusion.

\paragraph{Reduction Scheme for Multiple Models.}
To speed up the fusion process for multiple models and to limit the required memory, we introduce an iterative and a recursive version of NT, which we call \emph{NT-iterative} and \emph{NT-recursive} respectively. The iterative version iterates through the models and merges the current model with the next one until no model is left, i.e., similar to exponential weighted averaging. We expect a smaller accuracy drop post-fusion as a lot of weight is given to the last model of the iteration. The recursive variant recursively merges each half of the available models and then combines both. Both versions have the advantage that they only have to merge two models simultaneously while the latter is better parallelizable. 

\paragraph{Fusing Duplicate Models.}
A fundamental limitation of transplanting neurons is the case of fusing a model with itself, or a highly similar one, as the gained neurons are redundant and do not compensate for the pruned ones. We will showcase and analyze real-world scenarios where this can happen.

\section{Experiments}
In this section we conduct experiments on NT as a fusion method.
In section \ref{mainproperties} we analyze how Neuron Transplantation is performed in an effective way and how it fares in different settings. We deliberately prioritize demonstrating the feasibility and efficacy of our method over achieving the highest accuracy with the analyzed models on the chosen datasets. By focusing on validating the underlying principle of transplantation, we aim to provide a simple drop-in replacement for other more costly fusion methods. We therefore compare our method to state-of-the-art fusion techniques in section \ref{comparison}.

\subsection{Experimental settings}
\label{experimental_setting}

We first study NT's properties through ablation experiments and afterwards compare its performance against related model fusion approaches.
For the ablation studies, if not stated otherwise, we use an ensemble of two small neural networks with three hidden layers of width 512, referred to as  ``MLP'', and train it on the SVHN real-world image dataset~\cite{netzer} with the following hyperparameters: batch size of 256, 100 epochs, SGD-optimizer with a momentum of 0.9 and a constant learning rate of 0.01 and the cross entropy loss as the loss function.

For the comparison to other methods, we use the models LeNet~\cite{726791}, VGG11~\cite{Simonyan15} and Resnet18~\cite{DBLP:journals/corr/HeZRS15} and the additional image datasets MNIST~\cite{deng2012mnist}, CIFAR10 and CIFAR100~\cite{articlecifar}. LeNet and MNIST act as an easy classification task whereas CIFAR10/CIFAR100 with VGG11 or Resnet18 are slightly more challenging and allow a closer comparison to Optimal-Transport-fusion~\cite{singh2023model} as they used a similar setting. Since OT-fusion does not support biases and batch normalization, we remove them for these experiments. We make the following alterations to the hyperparameters to get a more accurate comparison to OT-fusion: we train the MLP and LeNet for 60 epochs (as the training process already converges) and VGG11/Resnet18 for 300 epochs. For VGG11 and Resnet18 we use a batch size of 128 and a learning rate of 0.05 with a decay of 0.5 every 30 epochs. 
\paragraph{Hardware and Software Environment.}
The experiments were conducted on a GPU4 node using 12 Intel Xeon Platinum 8368 CPUs and an NVIDIA A100-40 GPU with 40 GB memory. We trained our models using PyTorch~\cite{NEURIPS2019_9015}, a popular deep learning framework, and utilized CUDA~\cite{cuda}, NVIDIA's parallel computing platform, to leverage GPU acceleration for faster training. For our NT-fusion we utilize the pruning framework torch-pruning~\cite{fang2023depgraph} and for OT-fusion the Optimal-Transport solver POT~\cite{JMLR:v22:20-451}.

\subsection{Main Properties and Ablation Studies}
\label{mainproperties}

\begin{figure}
\includegraphics[width=0.5\linewidth]{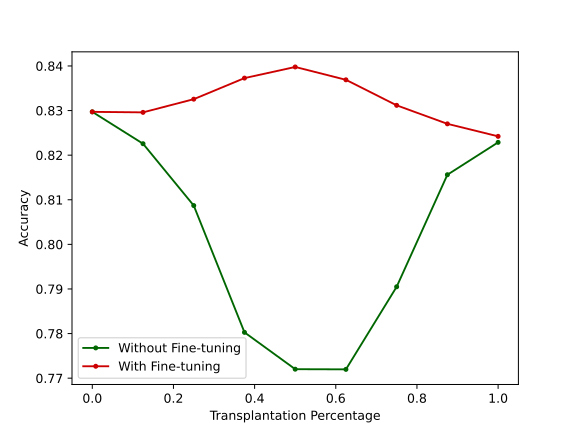}
\includegraphics[width=.5\linewidth]{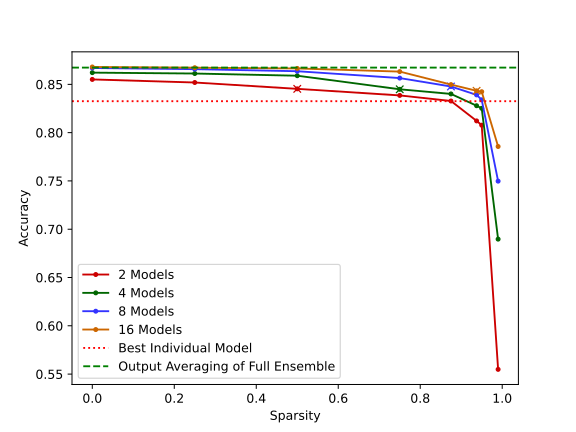}
\caption{Left: Transplanting different neuron amounts of one model into another. Without fine-tuning, test accuracy of the fused model drops symmetrically. With 3 epochs of fine-tuning, the fused model surpasses individual model performance peaking at a 50\% transplantation rate. Right: Fusing multiple models and pruning to specific sparsity ratios followed by 30 epochs of fine-tuning. Marked with ``x'' is the sparsity ratio, for which an individual model size is recovered. For more models, this shifts closer to a sparsity of one, where most performance is lost.
}
\label{fig:interpolation}
\end{figure}

Figure \ref{fig:interpolation} shows the key property of NT: At the cost of some current accuracy the potential to reach greater performance via fine-tuning is brought. The initial performance drop is a result of the pruning process. The loss of small-magnitude neurons is overcompensated by newly transplanted ones after fine-tuning.

\paragraph{Order of Operations.}
We consider the order of pruning, merging and potentially fine-tuning. We differentiate between three cases:
\begin{itemize}
    \item[1.] Prune-merge-fine-tune: The ensemble members are first pruned individually and then merged. This requires the least amount of memory and local pruning can be done in parallel.
    \item[2.] Merge-prune-fine-tune: The models are first merged and then jointly pruned. This has the advantage of considering all neurons in the pruning process.
    \item[3.] Merge-fine-tune-prune-fine-tune: The same as above with additional fine-tuning after merging, exploiting the potential of training the merged ensemble.
\end{itemize}

Table \ref{table:order_of_execution} shows that the additional fine-tuning step of option three does not pay off and that the second option, i.e., merging, then pruning,  slightly outperforms the first one at early epochs. We conclude that jointly selecting the neurons with the largest $L_2$-norm across all models is better than doing so locally, though the improvement is marginal at later epochs. We deliberately left out a possible fourth option: "Prune - Local Fine-tune - Merge". In this option, the ensemble members are pruned and fine-tuned individually and merged at the end. When doing this, it might be useful to ask what the purpose of the operation is. The merging in the last step decides whether this option is a simple ensemble-pruning technique or a model fusion technique. If the sole purpose is to prune the ensemble, then the merging should be omitted as it introduces cross-weights that are not required. If the purpose lies in model fusion, the required fine-tuning of all ensemble members needlessly increases the fusion time. Thus in both use cases, this option is sub-optimal.

\begin{table}[htbp]
\centering

\setlength{\tabcolsep}{2pt} 
\begin{tabular}{lcccccc}
\toprule
Order \textbackslash Fine-tuning Epoch &  4 & 7 & 10 & 13 & 16 & 19\\ 
\cmidrule{1-7}
Prune-Merge-Fine-tune & 53.87 & 54.11 & 54.16 & 54.64 & 54.57 & 54.83 \\ 
Merge-Prune-Fine-tune & $\textbf{54.14}$ & $\textbf{54.18}$ & $\textbf{54.43}$ & $\textbf{55.13}$ & 54.83 & 54.71 \\ 
Merge-Fine-tune-Prune-Fine-tune & 53.27 & 53.85  & 54.08 & 54.89 & \textbf{55.11} & \textbf{54.98} \\ 
\bottomrule
\end{tabular}

\caption{NT of two models with merging, pruning, and fine-tuning in different order. Fine-tuning is done for 20 epochs in total. The accuracies (in \%) are averaged over five different random seeds. First merging, then pruning  yields the best results for early epochs. At later epochs, all three methods have equal performance.}
\label{table:order_of_execution}
\end{table}

\paragraph{Multiple model fusion.} When merging an increasing amount of models, only large weights from each individual remain. The extreme case occurs when we merge $k$ models and prune to $k$ neurons in total. In that case we expect a saturation or loss of performance.
Figure \ref{fig:interpolation} (right side) shows the fusion behaviour for multiple models. While the ensemble continues to get better for a larger amount of models, the NT-fused model saturates.

In Table \ref{table:multiple_model_case} we compare the different reduction schemes to fuse multiple models. While all three methods succeed in extracting some ensemble performance, NT-iterative yields the worst results. We attribute this to the asymmetric weighting of the ensemble members. Both the baseline and hierarchical version perform similarly well.

\begin{table}[htbp]
\centering

\setlength{\tabcolsep}{2pt} 
\begin{tabular}{lcccccc}
\toprule
Method \textbackslash Models &  2 & 4 & 8 & 16\\ 
\cmidrule{1-5}
Ensemble & 84.61 & 85.85 & 86.43 & 86.82\\ 
Best Model & 82.57 & 82.92 & 83.18 & 83.19 \\ 
NT & 77.88 | 84.37 & 62.86 | \textbf{84.67} & 34.59 | \textbf{84.77} & 20.9 | 84.48\\
NT-Iterative  & 77.88 | 84.37 & \textbf{79.11} | 84.25  & \textbf{76.22} | 83.93 & \textbf{67.66} | 84.08\\
NT-Recursive & 77.88 | 84.37 & 61.98 | 84.65  & 46.82 | 84.72 & 42.35 | \textbf{84.63}\\
\bottomrule
\end{tabular}

\caption{NT and its variants fusing multiple models. 
We report the accuracy (in \%) in the format \textit{immediate$|$best}, which is after pruning and after fine-tuning for 20 epochs respectively, averaged over five seeds.
}
\label{table:multiple_model_case}
\end{table}
\paragraph{Width and depth.}
Tables \ref{fig:widths} and \ref{fig:depths} show how NT handles different layer widths and model depths. Smaller networks (in depth and width) profit slightly more from NT while all considered cases succeed in recovering some ensemble performance.
It is notable that the behaviour for the width is opposite to linear mode connectivity from Ainsworth et al. and from Singh et al. where they observe larger loss barriers at smaller widths and vice versa \cite{ainsworth2023git, singh2023model}. 

\begin{table}[htbp]
\centering

\setlength{\tabcolsep}{2pt} 
\begin{tabular}{lcccccccc}
\toprule
Method \textbackslash Width &  16 & 32 & 64 & 128 & 256 & 512 & 1024 & 2048\\ 
\cmidrule{1-9}
Ensemble & 68.26 & 76.66 & 81.40 & 83.99 & 84.29 & 84.93 & 85.12 & 86.07 \\ 
Best Model & 66.94 & 74.70 & 79.07 & 81.34 & 82.10 & 82.97 & 83.14 & 85.56\\ 
NT & 69.52 & 77.82  & 81.05 & 83.19 & 83.42 & 84.91 & 85.75 & 85.97\\ 
\bottomrule
\end{tabular}

\caption{Accuracies in \% on different layer widths of the MLP on the SVHN dataset. We merge two models trained for 100 epochs with a constant lr of 0.01. All training cases converged aside from $\text{width}=2048$ models. NT is usable for any considered layer width but seems to be especially useful for smaller widths (16 and 32).}
\label{fig:widths}
\end{table}

\begin{table}[htbp]
\centering
\centering

\setlength{\tabcolsep}{2pt} 
\begin{tabular}{lcccccc}
\toprule
Method \textbackslash Depth &  1 & 3 & 5 & 7 & 9 & 11\\ 
\cmidrule{1-7}
Ensemble & 82.76 & 84.78 & 84.17 & 82.83 & 82.03 & 79.44 \\ 
Best Model & 81.90 & 83.54 & 82.47 & 80.72 & 79.20 & 77.68\\ 
NT & 84.44 & 85.10  & 83.89 & 82.57 & 81.31 & 80.42\\ 
\bottomrule
\end{tabular}

\caption{Accuracies on different layer depths (in \%).  The ensemble consists of two models. Only the models for depths 9 and 11 did not converge after 100 epochs which is why further fine-tuning slightly increases the performance. Models of low depth seem to profit more from NT while in every case the best individual model's accuracy is surpassed.}
\label{fig:depths}
\end{table}

\paragraph{Failure case.}
Different to other fusion operations, NT applied on a copy of an individual model does not result in the identity operation. In this scenario, half of the neurons are discarded while the larger half is replicated. After training in step one, a single model has an accuracy of 83.02\%. Merging with itself and pruning leads to 67.63\%, and single performance is recovered after one epoch of fine-tuning.\\
A consequence of this property is the inefficiency to use NT as an alternative for averaging in synchronous SGD. When synchronizing every epoch with four models, the training process fails to train at all. It might be possible to circumvent the problem by diversifying the ensemble members, similar to Sun et al.~\cite{sun2017ensemblecompression}.

\subsection{Comparisons with Other Fusion Methods}
\label{comparison}
To compare to other fusion methods we report the accuracy after fusing, and after fine-tuning or distillation. We compare NT-fusion to vanilla averaging, and the state-of-the-art method of OT-fusion.

\paragraph{Hyperparameters for fine-tuning and distillation.}
We fine-tune or distill for 30 epochs, for the distillation we use a temperature of 2 and a soft target loss weight of 1 (discarding the hard targets) to train the fused-model with the ensemble teacher (outputs are averaged). For Resnet18 we also increase the number of epochs to 100, change the learning rate to 0.1 and the schedule to a 0.1-decay after epochs 50 and 80 (compared to the experimental setting in section \ref{experimental_setting}). 

\begin{figure}

\adjustbox{max width=\textwidth}{%

\begin{subfigure}{\textwidth}
  \centering
  \includegraphics[width=\linewidth]{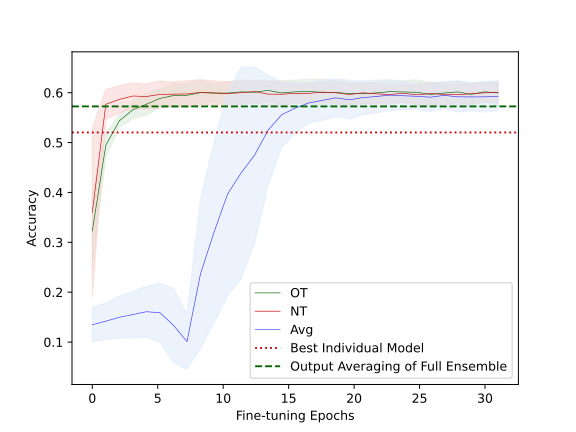}
  \label{fig:comparison_finetuning}
\end{subfigure}%

\begin{subfigure}{\textwidth}
  \centering
  \includegraphics[width=\linewidth]{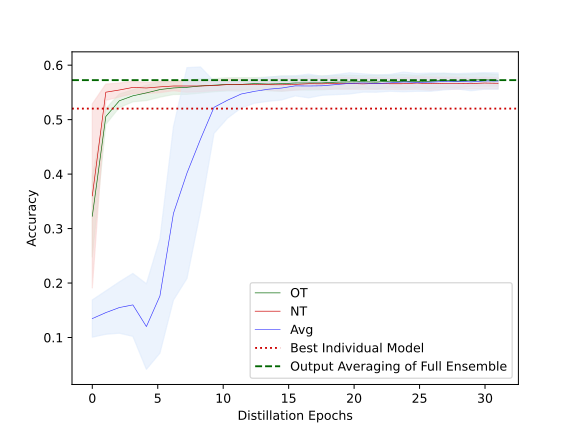}
  \label{fig:comparison_distillation}
\end{subfigure}
}

\caption{Mean accuracy plots for NT, OT and vanilla averaging for five different seeds after fusing two models. All methods beat individual accuracy after different amounts of fine-tuning (left) and distillation (right).}
\label{fig:comparison_vgg11_cifar100}
\end{figure}

\paragraph{VGG11 on CIFAR100.}
Figure \ref{fig:comparison_vgg11_cifar100} shows the mean accuracy of the fusion methods using fine-tuning and distillation. NT is quickest in converging, followed by OT and averaging. Vanilla averaging leads to a large drop in accuracy from which the model only recovers after many epochs of further training. In all cases individual performance is beaten, and performance of output averaging of the full ensemble is matched if not outperformed.

\paragraph{Various Datasets and Models.}
We further compare NT, OT and vanilla averaging on different model architectures and different datasets. Tables \ref{postfusion}, \ref{comparisontable_ft} and \ref{comparisontable_distill} show results for MLP, LeNet, VGG11 and Resnet18 on the datasets MNIST, CIFAR10, CIFAR100 and SVHN. Table \ref{postfusion} shows the post-fusion accuracies, Tables \ref{comparisontable_ft} and \ref{comparisontable_distill} the fine-tuning/distillation accuracies after 3 epochs and for the best checkpoint in 30 epochs. We do not use the large vision models VGG11 and Resnet18 for the MNIST and SVHN datasets. We make the following observations.

In most cases, NT has a larger accuracy immediately after fusing than OT. This indicates that the loss-barrier induced performance drop in OT (despite alignment) exceeds the drop caused by the exchange of the smaller neurons in NT. Without alignment, the performance drop is even larger as vanilla averaging has the least accuracy after fusion. These observations prevail for a few epochs of fine-tuning.\\
After about three to four epochs, the NT-fused model (and with slightly more epochs, the OT-fused model followed by the vanilla-averaged model) is on par with the best individual model outperforming it with further training.\\
All methods eventually reach a similar accuracy which lies above the best individual accuracy and below the ensemble accuracy with some exceptions where a better result is achieved. Any method can fail to do so in some cases, depending on the random seed.\\
Under distillation using soft targets, the choice of the student model only has a small effect on the outcome. Using the NT-fused model, the OT-fused model or even an individual model is viable while the vanilla-averaged model can still suffer from the loss barrier during the training process.

\begin{table}[htbp]

\centering

\begin{tabular}{lcc|ccc}
\toprule
\multicolumn{6}{c}{\textbf{Post-fusion}}\\
\midrule 
Model+Data & Ensemble & Best Model & Avg & OT & NT\\ 
\midrule 

MLP+CIFAR10 & $56.99 \pm 0.22$ & $54.91 \pm 0.35$ & $31.65 \pm 1.25$ & $\textbf{52.69} \pm \textbf{0.55}$ & $41.08 \pm 1.4$\\
\cmidrule{1-6}
MLP+MNIST & $98.22 \pm 0.07$ & $98.15 \pm 0.07$ & $72.1 \pm 11.84$ & $\textbf{96.91} \pm \textbf{0.12}$ & $94.19 \pm 1.99$\\
\cmidrule{1-6}
MLP+SVHN & $84.56 \pm 0.37$ & $83.16 \pm 0.35$ & $47.93 \pm 3.96$ & $78.99 \pm 0.64$ & $\textbf{80.26} \pm \textbf{1.27}$\\

\cmidrule{1-6}

LeNet+CIFAR10 & $65.96 \pm 0.61$ & $59.93 \pm 0.64$ & $15.62 \pm 2.37$ & $23.21 \pm 5.4$ & $\textbf{25.24} \pm \textbf{7.06}$\\
\cmidrule{1-6}
LeNet+MNIST & $99.16 \pm 0.04$ & $99.09 \pm 0.03$ & $57.34 \pm 6.56$ & $78.26 \pm 17.39$ & $\textbf{88.38} \pm \textbf{4.69}$\\
\cmidrule{1-6}
LeNet+SVHN & $89.23 \pm 0.34$ & $86.16 \pm 0.32$ & $27.34 \pm 2.49$ & $46.31 \pm 19.45$ & $\textbf{68.62} \pm \textbf{5.4}$\\

\cmidrule{1-6}

VGG11+CIFAR10 & $83.26 \pm 0.15$ & $81.49 \pm 0.28$ & $37.2 \pm 6.65$ & $\textbf{75.36} \pm \textbf{0.77}$ & $69.66 \pm 7.97$\\
\cmidrule{1-6}
VGG11+CIFAR100 & $57.25 \pm 1.07$ & $52.02 \pm 0.3$ & $13.48 \pm 3.82$ & $32.26 \pm 8.35$ & $\textbf{36.0} \pm \textbf{18.88}$\\
\cmidrule{1-6}

Resnet18+CIFAR10 & $87.79 \pm 0.25$ & $86.48 \pm 0.1$ & $\textbf{67.1} \pm 17.98$ & / & $57.72 \pm 34.31$\\
\cmidrule{1-6}
Resnet18+CIFAR100 & $69.72 \pm 0.65$ & $63.94 \pm 0.6$ & $\textbf{7.54} \pm 0.71$ & / & $1.06 \pm 0.06$\\
\bottomrule
\end{tabular}

\caption{Comparison of different fusion methods: Accuracies immediately after fusing multiple fully trained models with NT, OT and vanilla averaging. The results are averaged over 5 different random seeds.}
\label{postfusion}
\end{table}


\begin{table}[htbp]

\centering

\begin{tabular}{lcc|ccc}
\toprule

\multicolumn{6}{c}{\textbf{Fine-tuning}}\\
\midrule\\
Model+Data &  Ensemble & Epochs & Avg & OT & NT\\ 
\midrule 

\multirow{ 2}{*}{MLP+CIFAR10} & \multirow{ 2}{*}{$56.99 \pm 0.22$} & $3$ & $52.69 \pm 0.53$ & $53.31 \pm 0.46$ & $\textbf{53.6} \pm \textbf{0.46} $\\
& & $best$ & $54.57 \pm 0.82$ & $55.05  \pm 0.21$ & $\textbf{55.28} \pm \textbf{0.19}$\\
\cmidrule{1-6}
\multirow{ 2}{*}{MLP+MNIST} & \multirow{ 2}{*}{$98.22 \pm 0.07$} & $3$ & $97.8 \pm 0.07$ & $97.91 \pm 0.06$ & $\textbf{98.1} \pm \textbf{0.1}$\\
& & $best$ & $98.25 \pm 0.07$ & $\textbf{98.28} \pm \textbf{0.06}$ & $98.26 \pm 0.07$\\
\cmidrule{1-6}
\multirow{ 2}{*}{MLP+SVHN} & \multirow{ 2}{*}{$84.56 \pm 0.37$} & $3$ & $81.24 \pm 0.95$ & $82.2 \pm 0.8$ & $\textbf{84.22} \pm \textbf{0.42}$\\
& & $best$ & $83.97 \pm 0.2$ & $84.11 \pm 0.21$ & $\textbf{84.67} \pm \textbf{0.18}$\\

\cmidrule{1-6}

\multirow{ 2}{*}{LeNet+CIFAR10} & \multirow{ 2}{*}{$65.96 \pm 0.61$} & $3$ & $57.36 \pm 1.33$ & $\textbf{60.9} \pm \textbf{0.64}$ & $60.12 \pm 0.53$\\
& & $best$ & $62.88 \pm 0.49$ & $\textbf{63.03} \pm \textbf{0.42}$ & $62.0 \pm 0.45$ \\
\cmidrule{1-6}
\multirow{ 2}{*}{LeNet+MNIST} & \multirow{ 2}{*}{$99.16 \pm 0.04$} & $3$ & $98.69 \pm 0.21$ & $\textbf{98.73} \pm \textbf{0.26}$ & $98.68 \pm 0.42$\\
& & $best$ & $\textbf{99.14} \pm \textbf{0.05}$ & $99.11 \pm 0.07$ & $99.12 \pm 0.04$ \\
\cmidrule{1-6}
\multirow{ 2}{*}{LeNet+SVHN} & \multirow{ 2}{*}{$89.23 \pm 0.34$} & $3$ & $86.47 \pm 0.33$ & $86.6 \pm 0.23$ & $\textbf{87.39} \pm \textbf{0.37}$\\
& & $best$ & $87.99 \pm 0.28$ & $87.61 \pm 0.19$ & $\textbf{88.01} \pm \textbf{0.36}$ \\

\cmidrule{1-6}

\multirow{ 2}{*}{VGG11+CIFAR10} & \multirow{ 2}{*}{$83.26 \pm 0.15$} & $3$ & $78.66 \pm 1.14$ & $80.72 \pm 0.32$ & $\textbf{80.91} \pm \textbf{0.35}$\\
& & $best$ & $80.75 \pm 0.42$ & $81.08 \pm 0.29$ & $\textbf{81.12} \pm \textbf{0.27}$\\
\cmidrule{1-6}
\multirow{ 2}{*}{VGG11+CIFAR100} & \multirow{ 2}{*}{$57.25 \pm 1.07$} & $3$ & $15.52 \pm 5.19$ & $56.59 \pm 2.0$ & $\textbf{59.33} \pm \textbf{2.93}$\\
& & $best$ & $60.18 \pm 3.25$ & $\textbf{60.84} \pm \textbf{2.3}$ & $60.54 \pm 2.38$ \\
\cmidrule{1-6}

\multirow{ 2}{*}{Resnet18+CIFAR10} & \multirow{ 2}{*}{$87.79 \pm 0.25$} & $3$ & $70.57 \pm 7.19$ & / & $\textbf{78.74} \pm 1.55$\\
& & $best$ & $84.96 \pm 1.23$ & / & $\textbf{86.71} \pm 0.84$ \\
\cmidrule{1-6}
\multirow{ 2}{*}{Resnet18+CIFAR100} & \multirow{ 2}{*}{$69.72 \pm 0.65$} & $3$ & $56.0 \pm 2.09$ & / & $\textbf{58.51} \pm 0.99$\\
& & $best$ & $73.59 \pm 0.19$ & / & $\textbf{74.12} \pm 0.39$\\
\bottomrule

\end{tabular}

\caption{Comparison of different fusion methods: Accuracies in \% after fine-tuning for NT, OT and vanilla averaging. The results are averaged over 5 different random seeds.}
\label{comparisontable_ft}
\end{table}


\begin{table}[htbp]

\centering

\begin{tabular}{lcc|ccc}
\toprule

\multicolumn{6}{c}{\textbf{Distillation}}\\
\cmidrule{1-6} \\

Model+Data  & Epochs & Model0 & Avg & OT & NT\\ 
\midrule 

\multirow{ 2}{*}{MLP+CIFAR10} & $3$ & $52.79 \pm 0.16$ & $52.39 \pm 0.69$ & $\textbf{53.16} \pm \textbf{0.63}$ & $52.71 \pm 1.12$\\
& $best$ & $56.42 \pm 0.32$ & $56.42 \pm 0.32$ & $\textbf{56.77} \pm \textbf{0.13}$ & $56.62 \pm 0.25$ \\
\cmidrule{1-6}
\multirow{ 2}{*}{MLP+MNIST} & $3$ & $98.12 \pm 0.08$ & $97.75 \pm 0.23$ & $98.0 \pm 0.16$ & $\textbf{98.15} \pm \textbf{0.11}$\\
& $best$ & $98.28 \pm 0.08$ & $98.28 \pm 0.08$ & $\textbf{98.31} \pm \textbf{0.05}$ & $98.23 \pm 0.09$ \\
\cmidrule{1-6}
\multirow{ 2}{*}{MLP+SVHN} & $3$ & $83.89 \pm 0.15$ & $81.35 \pm 0.82$ & $83.17 \pm 0.26$ & $\textbf{84.45} \pm \textbf{0.35}$\\
& $best$ & $84.52 \pm 0.2$ & $84.52 \pm 0.2$ & $84.42 \pm 0.23$ & $\textbf{84.57} \pm \textbf{0.36}$ \\

\cmidrule{1-6}

\multirow{ 2}{*}{LeNet+CIFAR10} & $3$ & $61.06 \pm 0.59$ & $60.06 \pm 0.86$ & $\textbf{61.59} \pm \textbf{1.07}$ & $61.0 \pm 1.22$\\
& $best$ & $\textbf{64.23} \pm \textbf{0.36}$ & $\textbf{64.23} \pm \textbf{0.36}$ & $63.95 \pm 0.59$ & $63.51 \pm 0.37$ \\
\cmidrule{1-6}
\multirow{ 2}{*}{LeNet+MNIST} & $3$ & $\textbf{98.83} \pm \textbf{0.19}$ & $98.63 \pm 0.29$ & $98.77 \pm 0.09$ & $98.76 \pm 0.08$\\
& $best$ & $99.17 \pm 0.11$ & $99.17 \pm 0.11$ & $99.17 \pm 0.04$ & $\textbf{99.19} \pm \textbf{0.04}$ \\
\cmidrule{1-6}
\multirow{ 2}{*}{LeNet+SVHN}& $3$ & $87.02 \pm 0.2$ & $87.27 \pm 0.51$ & $87.3 \pm 0.21$ & $\textbf{87.95} \pm \textbf{0.45}$\\
& $best$ & $88.95 \pm 0.38$ & $88.95 \pm 0.38$ & $88.68 \pm 0.22$ & $\textbf{88.97} \pm \textbf{0.45}$ \\

\cmidrule{1-6}

\multirow{ 2}{*}{VGG11+CIFAR10} & $3$ & $\textbf{82.23} \pm \textbf{0.41}$ & $78.96 \pm 0.56$ & $80.89 \pm 0.33$ & $81.84 \pm 0.22$\\
& $best$ & $82.71 \pm 0.27$ & $82.71 \pm 0.27$ & $\textbf{82.9} \pm \textbf{0.33}$ & $82.54 \pm 0.28$ \\
\cmidrule{1-6}
\multirow{ 2}{*}{VGG11+CIFAR100} & $3$ & $55.35 \pm 0.67$ & $15.99 \pm 6.46$ & $54.37 \pm 1.23$ & $\textbf{55.91} \pm \textbf{1.26}$\\
& $best$ & $57.23 \pm 1.69$ & $57.23 \pm 1.69$ & $\textbf{57.39} \pm \textbf{1.28}$ & $56.84 \pm 1.2$ \\
\cmidrule{1-6}

\multirow{ 2}{*}{Resnet18+CIFAR10} & $3$ & $73.12 \pm 5.36$ & $67.85 \pm 5.93$ & / & $\textbf{73.46} \pm 5.37$\\
& $best$ & $85.48 \pm 0.75$ & $85.48 \pm 0.75$ & / & $\textbf{86.45} \pm 1.35$ \\
\cmidrule{1-6}
\multirow{ 2}{*}{Resnet18+CIFAR100} & $3$ & $\textbf{65.59} \pm 1.51$ & $59.04 \pm 1.62$ & / & $64.64 \pm 0.68$\\
& $best$ & $70.93 \pm 0.6$ & $70.93 \pm 0.6$ & / & $\textbf{71.14} \pm 0.68$ \\

\bottomrule

\end{tabular}

\caption{Comparison of different fusion methods: Accuracies in \% after distillation for NT, OT and vanilla averaging. The results are averaged over 5 different random seeds.}
\label{comparisontable_distill}
\end{table}

\paragraph{Memory and Time Measurements.}
We measure execution time and peak memory usage for OT, NT and averaging, varying the width of an ensemble consisting of two neural networks with one hidden layer. Table \ref{timememory} shows the execution time and RAM usage of the fusion (while the RAM is also influenced by storing the models and data batches). Note that the used GPU runs OOM for OT, which is the reason why we can only compare NT to vanilla averaging for larger layer widths. Averaging has a negligible memory usage, and wall time increases to a tenth of a second for the largest fusion. NT has similar negligible RAM usage but the time increases to $\sim$1s. This is still little compared to a normal training time of an ensemble which can be in the order of hours.

\begin{table}[htbp]
\centering

\setlength{\tabcolsep}{2pt} 

\begin{tabular}{lccccc}
\toprule
Method \textbackslash Width & 256 & 512 & 1024 & 2048 & 4096\\ 
\cmidrule{1-6}
\multirow{2}{*}{Averaging} & \textbf{0.002} s & \textbf{0.005} s & \textbf{0.015} s & \textbf{0.04} s &  \textbf{0.12} s\\
\cmidrule{2-6}
& \textbf{3.8} MB & \textbf{3.8} MB & \textbf{3.8} MB & \textbf{3.8} MB & \textbf{3.8} MB\\
\cmidrule{1-6}
\multirow{2}{*}{OT} & 0.73 s & 3.19 s & 14.57 s & / & /\\ 
\cmidrule{2-6}
& \textbf{3.8} MB & \textbf{3.8} MB & 22.1 GB & $>$ 38.67 GB & $>$ 38.67 GB \\ 
\cmidrule{1-6}
\multirow{2}{*}{NT} & 0.03 s  & 0.06 s &  0.15 s & 0.47s & 1.39 s\\ 
\cmidrule{2-6}
& \textbf{3.8} MB & \textbf{3.8} MB & \textbf{3.8} MB & \textbf{3.8} MB & \textbf{3.8} MB \\ 
\bottomrule

\end{tabular}

\caption{Execution time and RAM usage of different fusion methods. 
}
\label{timememory}
\end{table}



\section{Discussion and Conclusion}
The inference time and memory consumption of large pretrained ensembles can be large making it difficult to adapt them to downstream tasks with limited hardware. Distillation does not solve this issue as all ensemble members are present as the ensemble teacher during fine-tuning and recent alignment-based model fusion methods suffer from large memory requirements or search times.

In this work, we present a novel model fusion method called \textit{Neuron Transplantation}, that is able to fuse ensembles to a single model's size through computationally cheap pruning and little fine-tuning, retaining most of the ensemble performance. NT jointly prunes all models by removing neurons with the smallest $L_2$-norms and concatenates the remaining ones, setting up the fused model to retrain to a higher accuracy than any individual model. We theorize that only the large weights of the models are needed to set the fused model into a ``good'' loss neighborhood while the loss of the small weights can be compensated with further fine-tuning.

Compared to vanilla averaging, NT does not suffer from severe loss-barriers, and compared to OT, no costly permutation matrices need to be computed, enabling NT to fuse models with large widths. It can be used in combination with distillation to give a slight but cheap performance boost. Though our approach suffers from a saturation when fusing too many models (in our experiments >8) and from redundant neurons when the models are too similar, NT can be applied to various architectures like MLPs and CNNs. It's application to transformers~\cite{DBLP:journals/corr/VaswaniSPUJGKP17} and other use cases like federated learning or parallel SGD we leave for future work. 

We advocate NT as model fusion technique. Though its application as a pure ensemble-pruning technique also seems apparent, it is - by design as a canonical way of applying pruning to ensembles for model fusion - almost equivalent to pruning each ensemble member individually. Whether the joint pruning and joint training of NT can overcompensate for the newly introduced cross-weights remains to be seen. We also leave that for future work.

It might also be possible to heuristically decide whether the neurons are diverse enough for Neuron Transplantation to succeed. Such an approach could solve the information loss issue of using NT on really similar models.

\begin{credits}
\subsubsection{\ackname} 
This work is supported by the Helmholtz
Association Initiative and Networking Fund under the Helmholtz AI platform grant and the HAICORE@KIT partition.

\subsubsection{\discintname}
The authors have no competing interests to declare that are
relevant to the content of this article. 
\end{credits}







\bibliographystyle{splncs04}

\end{document}